\documentclass[twoside]{smgv}

\usepackage[T1]{fontenc} 
\usepackage[nocompress]{cite} 
\usepackage{amsmath}
\usepackage{multirow}
\usepackage{graphicx}

\hypersetup{pdffitwindow=true,pdfnewwindow=true,
plainpages=false,linktocpage=true,
breaklinks=true,
colorlinks=false,
pdftitle={Normal patch Retinex robust algorithm for white balancing in digital microscopy},
pdfauthor={Radosław Roszczyk, Artur Krupa, Izabella Antoniuk},
pdfsubject={Machine Graphics \& Vision 29(1/4):79-94, 2020},
pdfkeywords={auto white balance algorithm, microscope image processing, staining of microscopic slides, digital microscopy},
}

\autor{R.\ Roszczyk, A.\ Krupa, I.\ Antoniuk} 
\tytul{Normal patch Retinex robust alghoritm for white balancing in digital microscopy} 
\title{Normal Patch Retinex Robust Alghoritm\\for White Balancing in Digital Microscopy} 
\author{Radosław Roszczyk$^1$, Artur Krupa$^2$, Izabella Antoniuk$^2$\\
  {\footnotesize\sl $^1$Faculty of Electrical Engineering}\\
  {\footnotesize\sl Warsaw University of Technology, Warsaw, Poland}\\
  {\footnotesize\sl \href{mailto:radoslaw.roszczyk[at]pw.edu.pl}{radoslaw.roszczyk@pw.edu.pl}}\\
  {\footnotesize\sl $^2$Institute of Information Technology}\\
  {\footnotesize\sl Warsaw University of Life Sciences -- SGGW, Warsaw, Poland}\\
  {\footnotesize\sl \href{mailto:artur_krupa[at]sggw.edu.pl}{artur\_krupa@sggw.edu.pl}}
}

\begin{document}

\rok{2020}
\tom{29}
\numer{1/4}
\pierwszastrona{79}

\maketitle

\pagestyle{myheadings}
\thispagestyle{myfheadings}

\noindent

{\footnotesize{\bf Abstract.}\ \ The acquisition of accurately coloured, balanced images in an optical microscope can be a challenge even for experienced microscope operators. This article presents an entirely automatic mechanism for balancing the white level that allows the correction of the microscopic colour images adequately. The results of the algorithm have been confirmed experimentally on a set of two hundred microscopic images. The images contained scans of three microscopic specimens commonly used in pathomorphology. Also, the results achieved were compared with other commonly used white balance algorithms in digital photography. The algorithm applied in this work is more effective than the classical algorithms used in colour photography for microscopic images stained with hematoxylin-phloxine-saffron and for immunohistochemical staining images.}


\section{Introduction} 
\label{sec:intro}

The consistency of visible colours is one of the fascinating possibilities that the human eye provides. A person can look at an object from any angle, but regardless of the varying lighting conditions, the colour of the observed element will not change significantly. This effect is achievable due to the highly complex structure of the eye. In the case of humans, colour perception is compensated by the adaptive capacity of tissues and the capabilities of the human brain. However, computer algorithms cannot deal with this type of problem. At the same time, colour stability is the essential element in case of image processing and analysis as well as recognition of objects placed in processed scenes. Hence, colour stability plays a~significant role, especially in the algorithms of automatic image segmentation~\cite{Saha2015} or feature extraction~\cite{Thiran1996} in microscopic medical images.

The use of electronic image capture technology in medicine is based on the solution used in the past in conventional photography, i.e.\ the photosensitive film. The main difference, however, is the colour representation used in traditional and in microscopic photography.

In the case of digital photography, the image is created as a result of interaction of the incident light reflected from the surface of the photographed object with the light sensor. Recording of a~microscopic image (referred to as a~slide), for which the light source is placed centrally below the object on the laboratory glass, is realised by recording the flux passing through the object and falling on the optical system. As a result of this treatment, the image is more natural than in the case with the incident light and has a~white background. In this case, the light penetrates through the microscopic specimen.

The ability to adjust the white balance to the colour space is an important feature in the case of images obtained by optical sensors in modern electron microscopes. Most of the images are represented in the popular RGB space, so the resulting images are also stored in this space.

The eye and the physiological mechanisms which perform the image processing in the human vision system are not fully explored, but the eyes' ability to capture objects in the vicinity of the light beam which is reflected from them, is known. In such a~case the brain adjusts the input light spectrum so that the colour perception is consistent with the colour values of the observed object far from the beam. This, in turn, means that despite the different lighting parameters, the objects illuminated in this way remain perceptually uniform.

The research issues discussed are related primarily to microscopic images which are used by a very wide scientific community. Medical images, chemical reagents -- the use of microscopy allows researchers to see what cannot be seen with the human eye. The medical images constitute a~basis in the research methodology, and have a~direct impact on the therapy applied and its effectiveness.

Medical images require proper preparation both before their acquisition and after saving the digital form of the slide taken from the sample. A~comparison of two images made under the same laboratory conditions with the same staining may still show some colour discrepancy, as described in~\cite{Macenko2009}. One of the parameters which have the most important influence on the quality of the image is the white balance. An incorrectly performed optimization of the white balance of an image can affect the possibility of further processing of the used material. To our best knowledge, at present in the domain of medical imaging there is a~deficit of tools for white balance adjustment.

Unambiguity in medicine is critical, but it often happens that images are created by different medical groups, using completely different devices. Each team has a different approach to the calibration process, which is a condition of operational reliability. Just as the measurement of alcohol in the exhaled air should be performed with a certified and calibrated device, the materials used in the tests should be prepared in an appropriate manner. Despite the existence of various recommendations, the measurement conditions or rules describing the required sequence of actions, the images often remain without suitable colour preprocessing.

In the case of microscopic medical images, they are usually analysed by using feature extraction for objects present in such images. Inadequately prepared images can hinder the implementation of further operations. In recent years, many methods have been defined to solve this type of problem. In~\cite{Thiran1996} the extraction of features by the use of basic morphological segmentation and the description of the observed objects are shown, defining their geometric parameters along with their variability. However, the method reacts differently to the images illuminated with different intensity. Each time it is necessary to select the appropriate parameters of operation.

The colour information contained in the slides is very useful in the assessment of similarity of regions of the images. The availability of information in three colour channels instead of one grey level makes it possible to apply more advanced methods of analysis. Among such methods we can distinguish the L*a*b* segmentation method~\cite{Baldevbhai2012} or the automatic white balance methods~\cite{Garud2014}. We can also distinguish normalization methods, such as histogram extension, colour transfer method, or spectral methods~\cite{Hasna2016}. Mainly the latter is often used in microbiology and pathomorphology. It is based on estimating the tinting spectrum by adjusting the proportions of tinting to the intensity range for each pixel, even in the case of significant differences in shades.

\section{Images}
\label{sec:images}

The experimental part described in the paper was based on the use of microscopic medical images. The collection consisted of 200 images coming from microscopic scans of actual tissues. All the images are $1500\!\times\!1500$ pixels size. The images used were made using two types of staining which enhance important features necessary for medical analysis: hematoxylin-phloxine-saffron staining (HPS) and immunohistochemistry staining (IHC). Staining with IHC was performed with the application of two different biological markers: CK34 and KI67. Sets of images stained with IHC were collected using slides obtained from the Archives of the Military Medical University. The set stained with HPS was prepared from the OpenSlide public microscope slide collection (from~\cite{www_Mirax}, part of~\cite{www_OpenSlide}, described in~\cite{Goode2013}). Images were acquired using a variety of devices and in different optical and colour settings. Table~\ref{table:1} contains information about the number of samples and the file settings used during recording.

\begin{table} 
\caption{Description of data sets containing test images.\label{table:1}}
\centering
\begin{tabular}{c|c|c|c|c}
\textbf{Stain} & \textbf{Quantity} & \textbf{Series} & \textbf{Lens} & \textbf{Magnification}\\\hline
HPS & 100 & HPS & Hitachi HV F22CL & 20x\\
\multirow{2}{*}{IHC}
    & 50 & CK34 & CIS VCC-FC60FR19CL & 40x \\
    & 50 & KI68 &CIS VCC-FC60FR19CL & 20x \\
\end{tabular}
\end{table}

The images were divided into four series of 25 images for each type of staining. Each series was taken from different microscope slides, and the images in the series were selected randomly. This ensured that all the cases considered were independent and that the methods used were not closely correlated with the properties of the chosen data subset.

\section{Existing methods}
\label{sec:existing_methods}
Most of the widely available algorithms designed to ensure colour stability have been successfully implemented in colour photography~\cite{Lam2017}. However, such solutions have not yet been used for medical applications, and in particular not for the preparation of materials taken from microscopic sources.

The concept of white balance in digital technology is related to certain limitations of optical sensors performing the acquisition operations for the projection of the light beam reflected from the object onto the matrix area. White balance consists in adjusting the colour depending on the ambient light and the light falling onto the optics. Incorrect selection of light balance causes that the object correctly seen by the human eye will be, for example, too much inclined in the direction of {\em yellow} (giving the impression of warm) or too much biased versus to {\em blue} (giving the impression of cold).

\begin{figure}[b]
\centering
\includegraphics[width=.75\textwidth,keepaspectratio]{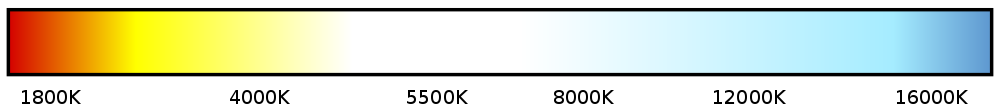}
\caption{Illustration of the light colour temperature scale indicating the relationship between the temperature (referenced to an ideal black-body radiator) and the colour of the light source perceived in a~given range -- warm for lower temperatures and cold for higher temperatures (saturation is amplified to make hues visible). Source:~\cite{www_ColorTemparatureGraphics}, used in~\cite{www_TemperaturaBarwowa}, among others. See also~\cite{www_ColorTemparature}.}
\label{fig:KelvinScaleBar}
\end{figure}

In its simplest form, this phenomenon is represented by the colour temperature scale (Fig.\ref{fig:KelvinScaleBar}). Leaving aside the physical issues of the nature of light, we can see that the determination of the temperature makes it possible to select appropriate parameters of the image colour transformation using this scale. It should be therefore determined whether a~given light is {\em cold} or {\em warm}. The higher the temperature on the scale, the colder the light, and warmer in the opposite direction. The aforementioned terms of {\em warm} and {\em cold} are theoretical concepts that characterise the generally accepted perception of colour by humans.

A typical home light source has a temperature of around 3000\usp{}K. Daylight (solar) also called {\em white} light has the temperature of around 5400\usp{}K during the day and 6700\usp{}K on a cloudy day. At night, it is almost completely blue, and the temperature ranges from 8000 to 10000\usp{}K.

In the case of photography or optics, it is usually necessary to calibrate the device to reflect the white correctly, and thus all other colours, based on the ambient light. In modern devices, there are often predefined profiles that offer the selection of temperature values in the average range. However, there may be cases when these values are far from the existing conditions, despite being set well.

\begin{figure}[tb]
   \centering
   \setlength{\unitlength}{\textwidth}
   \includegraphics[width=.85\textwidth,keepaspectratio]{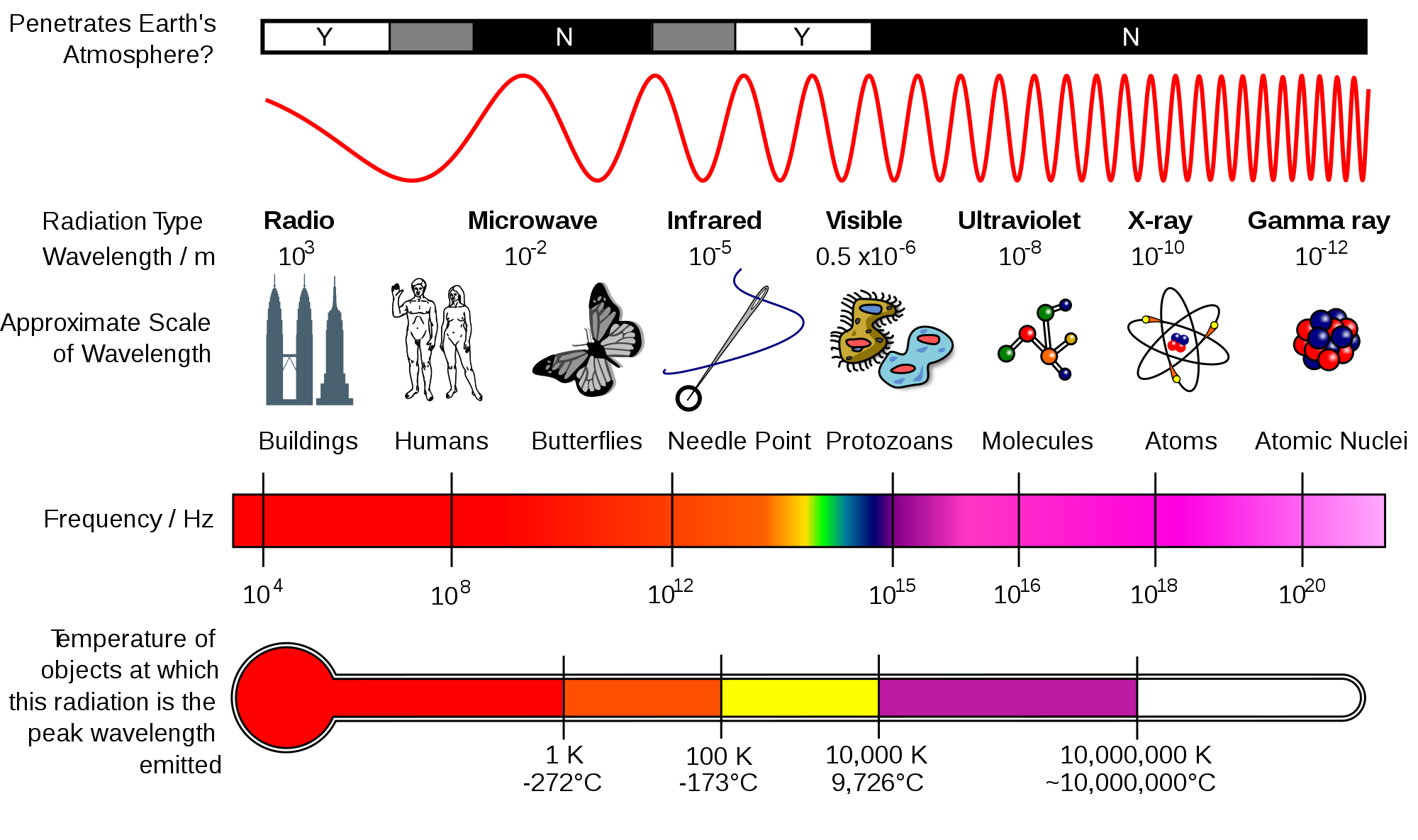}
   \caption{The spectrum of visible light, its location in the spectrum of electromagnetic waves and the information on which parts of this spectrum reach the earth surface. Source:~\cite{EMspectrum}.}
   \label{fig:widmo}
\end{figure}

In professional photography, the so-called grey cards, referring to surfaces reflecting 18\% of the light falling on them, are used. This solution was introduced and widespread by the Kodak company (card R-27~\cite{www_KodakR27}). The choice of the above value is not accidental. Each color can be defined by the parameters determining the electromagnetic wave's physical properties or by a subjective evaluation, a sensory representation related to the organ of vision. Human white light consists of a mixture of wavelengths ranging from $380$ to $790$ nanometers, and this is related to the spectrum of solar radiation reaching sea level~\cite{Biecek2016}. Different animals see different parts of the light spectrum and can use colour perception systems other than that of humans. In Figure~\ref{fig:widmo} the spectrum of the visible light and its location in the wider spectrum of electromagnetic waves is shown together with the information on which parts of this spectrum reach the earth surface. The spectrum of light emitted from a~surface depends on the spectrum of the incident light and on the physical features of the surface itself, which influence the light reflection. The relation of the light wave and the human perception of colour is influenced by the phenomenon of {\em colour metamerism}, which consists in that a~given colour impression can be received by various combinations of light wave lengths (for example, a~yellow colour is perceived when the yellow light is present and also when red and green lights are observed together). Therefore, the perceived colour of the surface can strongly depend on the type of the light source which illuminates it.

In the case of microscopic photography, we deal with relatively homogeneous illumination going from the source to the lens. In addition, the light stream is targeted and often has a~bounded region of incidence, which ensures that different microscopes produce similar images. However, this is not always the case, and similarly as for the cameras, it is recommended to calibrate the microscope before each measurement series. Such a~process is usually carried out by performing the colour correction for an image of a~clean glass and by selecting the appropriate settings based on the known parameters of the optics of the device.

The calibration process ensures that the measurements are comparable to each other over time; however, slight differences between devices introduce some uncertainty concerning the mutual similarity of the results. Thus, a~microscopic slide made with a~calibrated device from one manufacturer is not identical in colour to a~slide made with a~device of another one. In addition, the calibration procedure takes time that cannot be omitted in the case of regular measurements of a~large number of samples.

The idea behind this research was to bring about a~situation in which it is possible to collect images from the microscope without the necessity to carry out the calibration process, and without the necessity to limit the comparison of images to only images from the same device or the same type of the optical acquisition system used. Freeing oneself from these limitations became the basis for developing a~solution based precisely on the mechanism of white balance. Issues related to this have already been raised before, inter alia, in~\cite{Davis1931,Macenko2009,Hasna2016}.

We have founded our study on the \textit{retinex} theory\footnote{In the first papers the name of the theory and method, {\em retinex}, was spelled with lowercase first letter. In later publications the first letter became uppercase, like in the Retinex White Patch algorithm. So, we shall apply the lowercase and uppercase spellings in the respective fragments of the text.} originally introduced in~\cite{Land1971,Land1977,McCann1990}. This theory gave rise to the White Patch Retinex algorithm for enhancing the colour constancy~\cite{Land1977,McCann1990,Provenzi2008}. It underwent intensive development, see for example {\cite{Provenzi2005,Beltramio2009}} in which the \textit{retinex} theory was discussed. In this paper, the \textit{retinex} algorithm has been modified for colour correction of microscopic images.

\subsection{Retinex}
\label{sec:Retinex}

In the {\em retinex} theory the human impression of light intensity is treated as depending on the {\em relative difference} of image brightnesses rather than on the absolute values, which is based on extensive experimental material (described, among others, in~\cite{Land1964,Land1971,Land1977,McCann1990}, and many earlier works cited in~\cite{Land1964}). The term {\em lightness} is used instead of {\em brightness} or {\em intensity}.  In the simplest form of the {\em retinex} algorithm~\cite{Land1971} all the random paths leading from a~random point in the image to the specified point, in which the lightness value is calculated (the term {\em lightness} is used in the literature of the {\em retinex} theory instead of {\em brightness}). In the first version of the algorithm the relative value of the lightness resulted from the comparison of brightnesses on the individual paths with the value of the lightness of a~specified pixel. Considering all the paths is computationally complex but makes it possible to perform a~full analysis of an image. The result is the average of the quotients of values of all the subsequent lightness value changes along the paths. It is described by the so calculated lightness value $L(x)$ over all the paths according to the formula (we shall use a~clear description of the {\em retinex} algorithm from~\cite{petro2014multiscale}):
\begin{eqnarray}
        L(x) &=& \frac{ \sum_{k=1}^{N} L(x; y_{k}) }{ N } \; \label{eqn:Lx}
\end{eqnarray}
where: $N$ -- number of all the paths, $x$ -- starting point of a~path, $y_{x}$ -- final pixel of each path, $L(x;y_{k})$ -- relative value of pixel lightness for a single path:
\begin{eqnarray}
        L(x; y_{k}) &=& \sum_{t_{k}=1}^{n_{k}} \delta \left[ \log\frac{I(x_{t_{k}})}{I(x_{t_{k+1}})} \right] \; \label{eqn:Lxyk}
\end{eqnarray}
where: $n_{k}$ -- number of pixels in a~single path, $t_{k}$ -- subsequent iterated index of pixels in the range, $x_{t_{k}}$ -- lightness in a~current pixel, $x_{t_{k+1}}$ -- lightness value in the next pixel, $\delta$ -- threshold of contrast for the given $t$, where $t \in [0,1]$:
\begin{eqnarray}
    \delta(s) &=& \begin{cases}
        s & \text{if } |s| \ge t \\
        0 & \text{if } |s| < t
        \end{cases} \label{eqn:thresh}
\end{eqnarray}

The idea of the algorithm is to find the largest value along the path. In the case of an analysis path by path, the reset system sets to zero the previously found value, if a~new value is greater than the one found previously, so the new value becomes the largest one. Additionally, the algorithm performs the task of assuring that its start takes place in the region where the largest lightness value appears. The details of that concept were described in~\cite{Provenzi2005}.

A relatively important modification of {\em retinex}, applied in this research, was the algorithm of Single-Scale Retinex (SSR)~\cite{jobson1997properties} (submitted in 1995). It is based on the classic choice of a~typical local value of lightness with the {\em nearest neighbours} method (NN). Is is crucial to take into account each of the channels of the colour model in this process. From the point of view of the efficiency and ease of description of the phenomenon, the HSV model should be more suitable for the {\em retinex} algorithm than the RGB model. This is related above all with the search for the lightness path, which is directly the $V$ component (Value) in the HSV model. For the imaging task and for the analysis of a~typical image the RGB model is used, however, because the lightnesses in the separate hue channels are considered in {\em retinex}, according to the existence of separate receptors for the long, middle and short wavelengths in the human visual system.

The general form of {\em retinex} in a given point for one iteration is described with the formula
\begin{eqnarray}
        R_{i}(x,y) &=& \log(I_{i}(x,y)) - \log(I_{i}(x,y) * F(x,y)) \label{eqn:SSR}
\end{eqnarray}
where: $I_{i}$ -- input image for one channel of the source ($i-tego$), $F$ -- normalised function of the neighbourhood for the pixels belonging to this neighbourhood.

The function $F(x,y)$ proposed by the author of the algorithm is the classic cross averaging method
\begin{eqnarray}
        F(x,y) &=& \frac{C}{x^{2} + y^{2}} \label{eqn:FunF}
\end{eqnarray}
where $C$ -- normalization coefficient.

Alternatively, the Gaussian function is used in SSR:
\begin{eqnarray}
        F(x,y) &=& C * exp^{\frac{-(x^{2} + y^{2})}{2\sigma^{2}}} \label{eqn:GauF}
\end{eqnarray}
where: $\sigma$ -- scale of the filter (deviation). According to the experiments described in~\cite{jobson1997properties}, $\sigma = 80$ is a~good value for calculations. The use of the convolution operation applied before calculating the logarithm in~(\ref{eqn:SSR}) has also been demonstrated.

\subsection{White Patch}

White Patch is the method based on the {\em Retinex} theory, which assumes the full use of the possibilities of the active areas of the eye (the rods), which capture the complete information coming from the light falling on them. The brightest point is the one that reflects 100\% of the light from the chosen colour~\cite{Garud2014}.

Taking into account that the input image is most often described in the tri-colour component (RGB), the operation should be performed separately for each component. Due to the fact that all the rods are responsible for the white colour, therefore, the obtained range of stimuli is maximised to the entire spectrum by proportionally changing the values. The method is then based on adopting the following transformed data format:
\begin{eqnarray}
        R_\mathrm{max} = \max_{x,y} R(x,y) \; ,\\
        G_\mathrm{max} = \max_{x,y} G(x,y) \; ,\\
        B_\mathrm{max} = \max_{x,y} B(x,y) \; .
\end{eqnarray}
where: $(x,y)$ -- coordinates of the point with maximum value lightness of the given colour component, respectively, red ($R$), green ($G$) or blue ($B$).

The White Patch method has been crossed out for digital photography due to the lack of unequivocal effectiveness for both grayscale and colour scale images. However, microscopic images in the processing are ultimately converted to grayscale, so in this case, it is possible to implement a combination of these methods.

The White Patch method for the path mechanism is realised in an the way analogical to the {\em retinex} algorithm. For the image with high resolution and compression ratio, and with the necessity of representing the hue in wider ranges, the Multi-Scale Retinex (MSR) has been introduced~\cite{Jobson1996}. It is a~solution founded on weighted summing of single results for each path separately:
\begin{eqnarray}
        R_{\mathrm{MSR}_{i}} &=& \sum_{n=1}^{N} \omega_{n}R_{n_{i}} \; \label{eqn:MSRn} ,\\
        R_{n_{i}} &=& \log(I_{i}(x,y)) - \log(I_{i}(x,y) * F(x,y)) \label{eqn:MSRF} \; ,
\end{eqnarray}
where: $N$ -- scale of the solution with respect to the coefficient $\sigma$ (number of SSR components), $\omega$ -- weight for each scale.

\section{Normal Patch Retinex}
\label{sec:NormalPatchRetinex}

Taking into account the above information, an algorithm under the name \textit{Normal Patch Retinex} (NPR) is proposed that uses the advantages of the White Patch and Retinex algorithm, based on normalisation using the luminance values of both base algorithms and chrominance matching to obtain an optimised base algorithm. The preparation method is based on the modification of the above-mentioned algorithms.

\begin{figure} 
\centering
\includegraphics[width=0.55\textwidth,keepaspectratio]{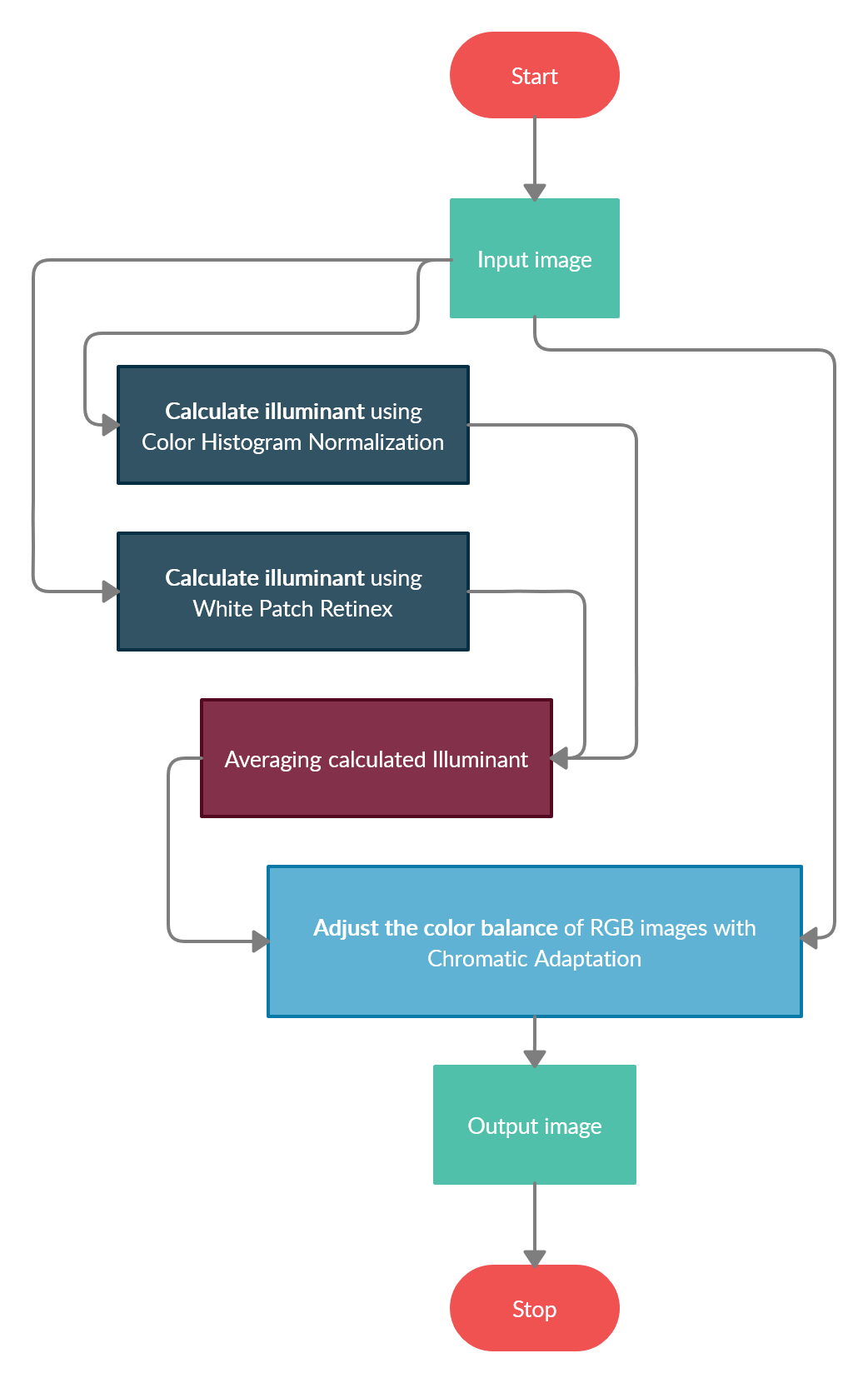}
\caption{Diagram representing the general outline of the Normal Patch Retinex algorithm.}
\label{fig:Flow}
\end{figure}

To understand the procedure, one can visualise the sequential operations according to the scheme shown in Fig.~\ref{fig:Flow}.

The source image is read in at the beginning. In the second step, this image is processed with the classic histogram normalization, to find its corrected brightness. In the third step, which can be performed in the parallel way with the second step, the original input image is processed with the White Patch Retinex method. The images from the second and third step are stored separately. These average value of these two images, pixel by pixel, are calculated and stored as the {\em Average Illuminant}.

The original, source image is then transformed by colour balance adjustment, with the use of the {\em Average Illuminant} image. The image obtained in this way is stored as the result of the white balance equalization process.

This algorithm was implemented, on the basis of the results received in the trials with the individual algorithms an in the combined methods, presented in Table~\ref{table:2}.

\begin{table} 
\caption{Average values of the angular error for the input images of each set for the applied white balance methods.\label{table:2}}
\centering
\begin{tabular}{c|c|c}
\textbf{Method} & \textbf{IHC} & \textbf{HPS}\\\hline
Original & 0.48 & 1.65 \\
Mean Shift Gray Pixel & 0.98 & 4.14 \\
Color Histogram Normalisation & 1.92 & 2.22 \\
Gray World & 0.84 & 5.30 \\
Cheng's Principal Component Analysis & 1.73 & 3.25 \\
White Patch Retinex & \textbf{0.47} & 1.63 \\
All Gray Pixels & 2.71 & 1.64 \\
YUV Gray Pixels & 0.73 & 3.59 \\
Normal Patch Retinex & 1.07 & \textbf{1.37} \\
\end{tabular}
\end{table}

The received results indicate univocally that the White Patch Retinex algorithm and the Normal Patch Retinex proposed in this paper yield the most profitable results for the set of the tested images, from the viewpoint of the white balance. The IHC and HPS staining was considered here.

If the whole available set of images and the unification of the method for each type of images acquired with the microscopic method, the proposed Normal Patch Retinex algorithm clearly appears as the most efficient one (Table~\ref{table:3}).

\begin{table} 
\caption{Standard deviation and averages of the angular error for all input images.\label{table:3}}
\centering
\begin{tabular}{c|c|c}
\textbf{Method} & \textbf{Angular Error} & \textbf{Standard deviation}\\\hline
Original & 1.87 & 1.66 \\
Mean Shift Gray Pixel & 2.30 & 1.93 \\
Color Histogram Normalisation & 1.95 & 0.84 \\
Gray World & 2.76 & 2.09 \\
Cheng's Principal Component Analysis & 2.11 & 1.26 \\
White Patch Retinex & 1.86 & 1.67 \\
All Gray Pixels & 2.55 & 3.60 \\
YUV Gray Pixels & 1.67 & 1.58 \\
Normal Patch Retinex & \textbf{1.32} & \textbf{0.77} \\
\end{tabular}
\end{table}

\begin{figure}
\centering
\includegraphics[width=0.73\textwidth,keepaspectratio]{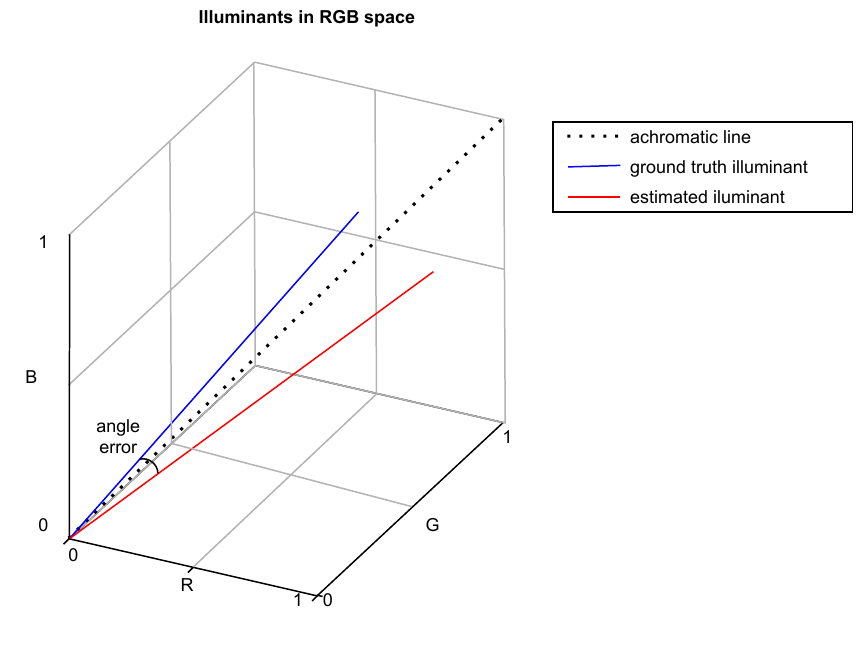}
\caption{Luminance lines in the colour space depending on the RGB parameter and the estimated luminance value converted from the reference value.}
\label{fig:Illuminants}
\end{figure}

Angular error is a helpful metric to evaluate the estimation of an illuminant against the ground truth. The smaller the angle between the illumination determined for the ground truth and the estimated illumination, then the better the quality of the estimate. To better understand how to use the determined luminance value, make assumptions as illustrated in Fig.\ref{fig:Illuminants}.

\section{Results}
\label{sec:results}

\begin{figure} 
 \centering
 \begin{minipage}[b]{0.49\textwidth}
  \includegraphics[width=\textwidth]{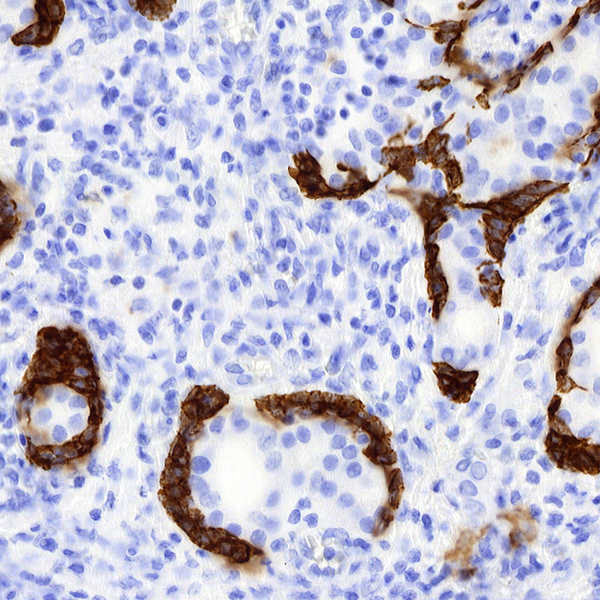}
 \end{minipage}
 \hfill
 \begin{minipage}[b]{0.49\textwidth}
  \includegraphics[width=\textwidth]{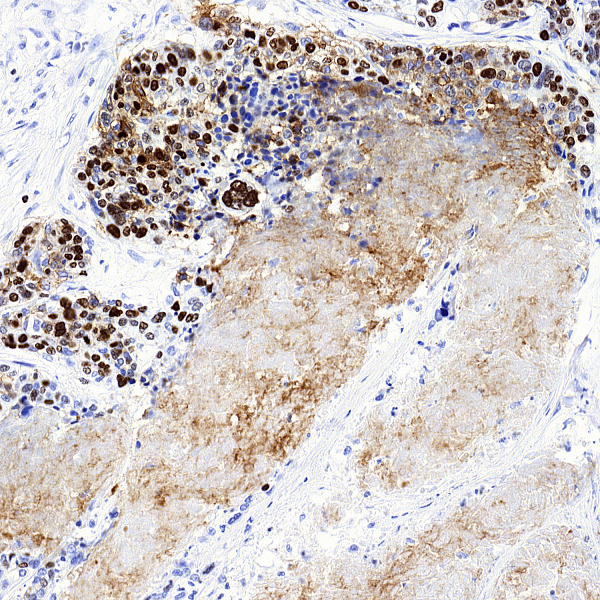}
 \end{minipage}
 \makebox[0.49\textwidth][c]{\bf a}\hfill\makebox[0.49\textwidth][c]{\bf b}
 \caption{%
   Comparison of two slides made under different laboratory conditions with different staining methods. IHC samples, stained with:
   ({\bf a})~CK34,
   ({\bf b})~KI67.
 }
 \label{fig:SlideCompare}
\end{figure}

Images from the slide sets described in Section~\ref{sec:images} were used for testing (sample images are shown in Fig.~\ref{fig:SlideCompare}). The images were mixed between sets to verify the effectiveness of the adapted algorithm.

In the present study, the original form of staining does not matter much for the algorithms used, due to that the images are transformed to the grayscale.

For research and comparison purposes, the methods used mainly in digital photography to balance white in input images were used here. The summary can be found in Table~\ref{table:2}. As it can be seen, the most appropriate white balance coefficients were achieved for three methods: the aforementioned retinex method, the method based on the White Patch, and the grayscale method for the HSV colour space.

Taking into account the above results, it was assumed that the combination of the mentioned methods might be a solution that makes further use of medical images independent from the parameter of the amount of light or method used for microscopic data acquisition. The first step is to convert the input image to grayscale.

The Table~\ref{table:3} contains the summary results of the comparison of the proposed algorithm in comparison with popular algorithms used in colour photography. For each value of the angular error, standard deviations were calculated for the entire study population consisting of 200 input images. The result achieved by the Normal Patch Retinex algorithm is by far the best with the smallest deviation of the results of the tested samples from the mean value.

In the case of implementing the solution for microscopic images, it was assumed that the converted grayscale image should be normalised. The transfer of RGB colours to the grayscale space ensures that the components retain their values despite the expansion of the colour spectrum.

In traditional photography, algorithms search for the darkest and brightest places to determine the range of values across the entire set of spaces. To extend the histogram for a grey image in a similar way, it is necessary to find pixels with dark and light values, respectively. The difference of this algorithm for microscope slides is that the light surfaces are the passage of light through the object, and the dark ones – places where the beam was stopped.

In other words, while in photography, the light falling on a bright object reflects off it and hits the sensor in the lens, in the case of a microscopic object, the reflected light does not reach the lens because it reflects off the glass. The algorithm should, therefore perform the colour assignment in the opposite way than the value of the scanned slide indicates.

\begin{figure} 
\centering
\includegraphics[width=1.00\textwidth,keepaspectratio]{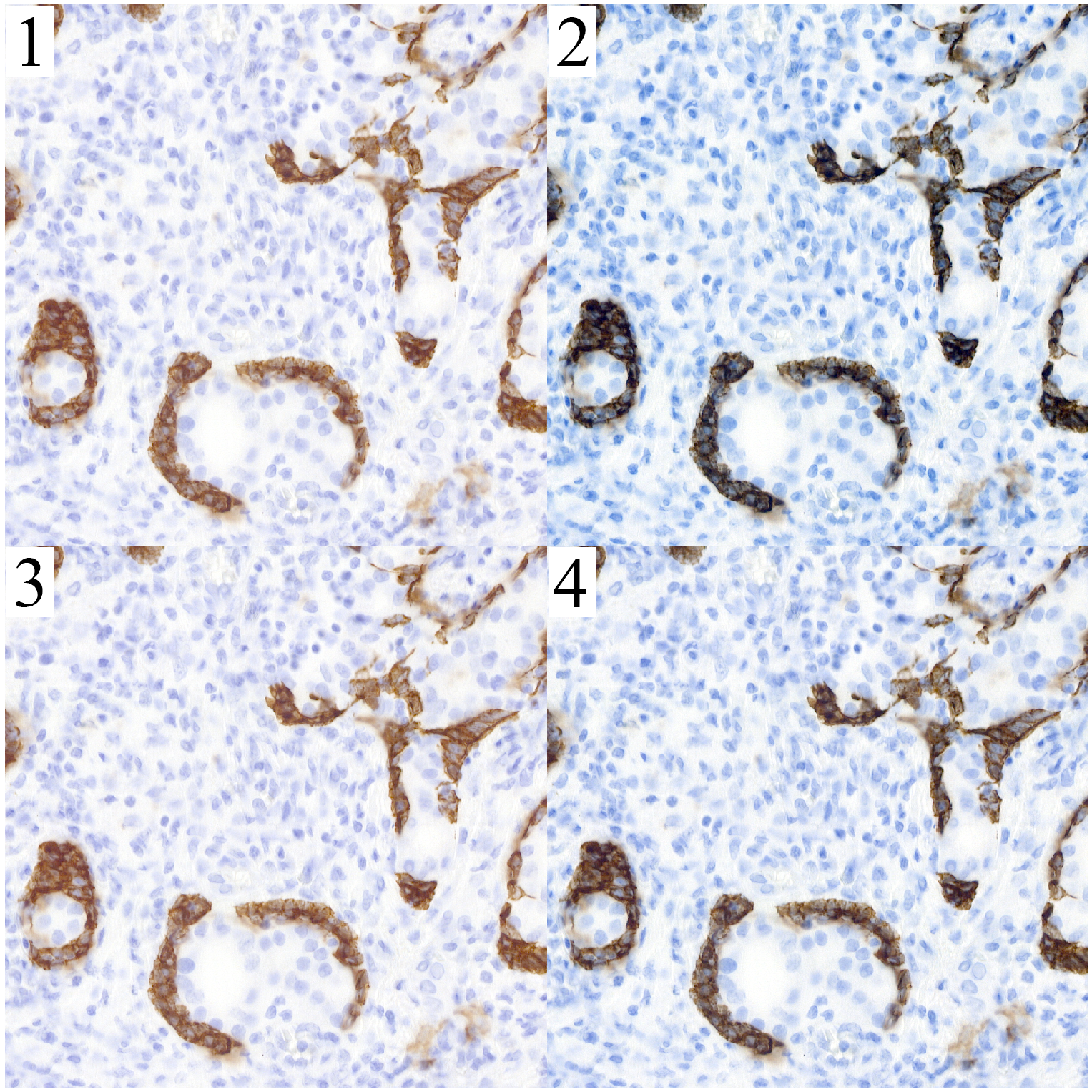}
\caption{Comparison of the effects of white correction on the basis of a preparation with immunohistochemistry staining. Algorithm numbering:
({\bf 1})~original image,
({\bf 2})~white balancing with Colour Histogram Normalisation,
({\bf 3})~white balancing with White Patch Retinex,
({\bf 4})~white balancing with Normal Patch Retinex.}
\label{fig:WPR}
\end{figure}

Referring to the microscopic output image constructed in this way, the result is an image with the most optimally matched colour balance. However, to verify the thesis, the Normal Patch Retinex needs to be implemented and carried out for the collected data set. Fig.~\ref{fig:WPR} shows sequentially numbered results of the algorithm's operation.

According to the aforesaid method of calculating the luminance in the RGB space, the correct determination of the estimated value requires to average the value calculated by the algorithm and the reference luminance value. The difference between the offset angles of both luminance is the value sought for which the white balance method using chrominance adaptation can be later used.

\section{Conclusion}
\label{sec:conclusion}

The proposed white balance correction algorithm in microscopic images allows for quick and effective colour equalisation. This algorithm does not require the prior preparation of input data or other pre-processing methods. The big advantage of the Normal Patch Retinex algorithm is its speed, full automaticity and ease of use.

The presented algorithm solves the problem of white balance equalization in a way dedicated to microscopic imaging. Previously, the algorithms used in colour photography were used for medical imaging. This algorithm properly corrects the white balance in images of tissues stained with different methods. The algorithm can be successfully used in the process of pre-treatment of single scans of microscopic slides or the entire series of microscopic images. The use of NPR to align the colour space of a series of images allows obtaining a consistent colour space for all processed images.


\end{document}